\documentclass[conference]{IEEEtran}
\IEEEoverridecommandlockouts
\usepackage{cite}
\usepackage{amsmath,amssymb,amsfonts}
\usepackage{algorithmic}
\usepackage{graphicx}
\usepackage{textcomp}
\usepackage{xcolor}
\def\BibTeX{{\rm B\kern-.05em{\sc i\kern-.025em b}\kern-.08em
    T\kern-.1667em\lower.7ex\hbox{E}\kern-.125emX}}
\begin{document}

\title{Detecting Deepfakes with Metric Learning}
% Fighting Deepfakes on Social Media platforms: Are we doing enough?

% Weak-Supervised Adaptation to Detect Deepfakes in High Resolution Videos


\author{
\IEEEauthorblockN{Akash Kumar}
\IEEEauthorblockA{\textit{MANAS Lab} \\
\textit{IIT Mandi}\\
India \\
akash\_bt2k15@dtu.ac.in}
\and
\IEEEauthorblockN{Arnav Bhavsar}
\IEEEauthorblockA{\textit{MANAS Lab} \\
\textit{IIT Mandi}\\
India \\
arnav@iitmandi.ac.in}

% **Annonymous Author(s)

}

\maketitle

\begin{abstract}
With the arrival of several face-swapping applications such as  \textit{FaceApp, SnapChat,  MixBooth, FaceBlender} and many more, the authenticity of digital media content is hanging on a very loose thread. On social media platforms, videos are widely circulated often at a high compression factor. In this work, we analyze several deep learning approaches in the context of deepfakes classification in high compression scenarios and demonstrate that a proposed approach based on metric learning can be very effective in performing such a classification. Using less number of frames per video to assess its realism, the metric learning approach using a triplet network architecture proves to be fruitful. It learns to enhance the feature space distance between the cluster of real and fake videos embedding vectors. We validated our approaches on two datasets to analyze the behavior in different environments. We achieved a state-of-the-art AUC score of 99.2\% on the Celeb-DF dataset and accuracy of 90.71\% on a highly compressed Neural Texture dataset. Our approach is especially helpful on social media platforms where data compression is inevitable. 

\end{abstract}


\section{Introduction}
With the rapid increase of online streaming platforms, there is a dire need to check the authenticity of the videos. In Youtube alone, 300 hours of videos are uploaded every minute. On a daily basis, 5 billion videos are watched and 1 billion hours are streamed, that's Facebook and Netflix streaming combined. The rise of deepfakes in recent years seriously raises concerns about the authenticity 
of digital content by media and other online streaming platforms. Generative architectures are excellent for aiding in boosting the performance of deep learning architectures by satisfying the need for large datasets, and in general to explore the creative power of deep learning. However, such approaches have also resulted in Deepfakes, which are now been utilized for nefarious purposes to manipulate the images of politicians, famous actors, etc. Many politicians and actors are becoming victims of Deepfakes. For criminal purposes, forensic videos are altered using novel methods such as \textit{faceswap} and \textit{faceswap-GAN}. 

Various applications are using human faces to transform it into sophisticated fun images like modifying the age, changing the gender, etc. In exchange, the users are giving away their face data to these companies, which can further be used for wrongful purposes. 

When the manipulated videos are shared on social apps, their quality is reduced to make it convenient for uploading and downloading through those apps. In high-quality videos, a small amount of fuzziness around the face warping can be visualized. However, in low-quality videos, users are unable to detect distinguish whether the videos are authentic or fake and the videos are forwarded to large groups of people. Such manipulations can have large scale effects, from politics to the entertainment industry.  
For instance, videos of politicians participating in certain events or announcing a public service, that never really existed sabotage the image of the politician. Similarly, fake porn videos of actors commonly circulated online. 

\begin{figure}[t]
\centering
\includegraphics[width=\columnwidth]{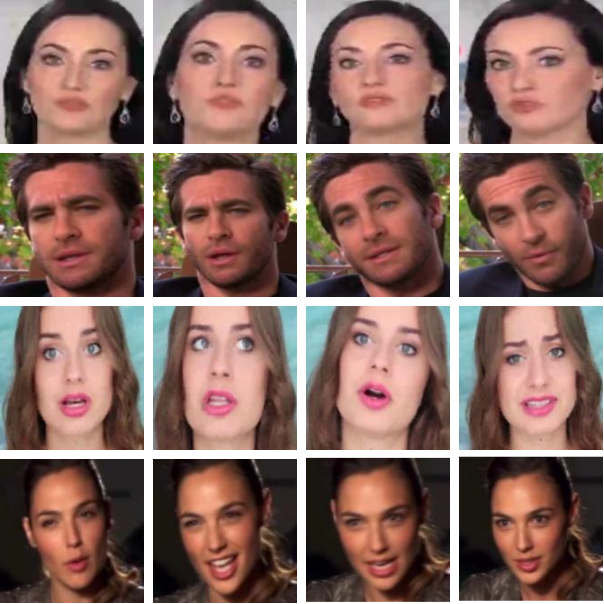} % Reduce the figure size so that it is slightly narrower than the column. Don't use precise values for figure width.This setup will avoid overfull boxes. 
\caption{The first two rows depicts the deepfake frames of FF++ and CelebDF dataset. Next two rows are examples of original sequences of the respective datasets.}
\label{fig3}
\end{figure}

To counter (detect) such as video manipulation, several algorithms using handcrafted features, deep learning algorithms, and lately GAN-based methods are being explored. For instance, handcrafted approaches involve methods for steganalysis, detecting 3D head pose inconsistencies, etc. Several such existing approaches are summarized in \cite{survey_2} and \cite{survey_1}. However, there is still scope of improvement over the state-of-the-art for detecting deepfakes, especially on challenging data such as the Face Forensics (FF++) dataset \cite{faceforensics}.

\textbf{Contributions} In this paper, our main contributions are as follows: 1) Improve the binary classification of Deepfakes on new second-generation Celeb-DF\cite{celebdf} dataset and the FF++ dataset \cite{faceforensics}, using a metric learning approach. 2) Experimentally demonstrating the performance of various contemporary methods and their variations to classify videos in high compression factors on the FF++ dataset. 3) Boosting the accuracy in low-resolution videos with minimal dependency on the number of frames required per video, as compared to recent existing methods.

The paper is arranged a follows: Section II discusses about the existing approaches in deepfakes video generation, datasets available and classification approaches. Section III goes through the approaches we used to improvise the video classification in high compression factors. After that, we discuss our experimental results in Section IV. Further, future directions our work is discussed in Section V.

%-------------------------------------------------------------------------
\section{Related Work}

\subsection{Deepfakes Generation methods}
With the rise of engagement on social media platforms, many applications are now based on face-swapping technologies. \cite{face2face, fsgan, neuraltextures, df_creation_detection} Many novel approaches have been introduced in recent years. Thies \textit{et al.} in his work Face2Face\cite{face2face} produces a real-time \textit{facial reenactment} video. Initially, the training corpus contains the identity of the target actors. Then, they track the expressions of both the source and target candidate and, smoothly apply a deformation to transfer the expression. After that, the face is reconstructed. Compared to other depth-based methods, they have used tracking across frames and appearance information in RGB videos. Deepfakes \textit{faceswap} uses an encoder-decoder based approach to alter the face of the target candidate. Two same encoders with shared parameters for both the candidates, after that, the decoder for target candidate is used to transfer the facial expressions of source candidate. \textit{faceswap-GAN} incorporate adversarial and perceptual loss to further improve facial reconstruction. In Neural Texture synthesis\cite{neuraltextures}, 3D reconstruction of images is done in imperfect geometry conditions and produced at real-time rates. High-level encoding of surface appearance and the 3D environment is captured. It helps the network to easily manipulate the re-renderings of the environment of source candidate on target candidate. Nirkin \textit{et al.} \cite{fsgan} proposed a method that does not even need the target actor to be present inside the training corpus. Moreover, they used two new loss functions stepwise consistency loss and blending loss. Stepwise consistency loss regulates the transfer of source candidate's appearance to target candidate and reconstruction loss keeps in check of the realism of face reconstruction. Poisson blending helps to blend in the two faces seamlessly keeping in view with the background environment.

\subsection{Deepfake Video Datasets}
Many new large and realistic Deepfake Video datasets start coming up from the year 2018, as the use of GANs in several research directions started blooming.\cite{survey_2} The Deepfake-TIMIT was the first dataset that synthesized videos using faceswap-GAN to generate 640 videos. Videos were divided into low and high quality depending upon the resolution of images that were 64 and 128 respectively. Fake Face in the Wild dataset, involves splicing and deepfake approaches to generate only 150 videos of frame sizes ranging from 480p to 1080p. A large and diverse dataset consisting of numerous manipulations for automatically generating faces, FF++ was released last year. It contains 1000 real videos carefully extracted from the Youtube-8M dataset. Then, four types of approaches were employed to re-render the expression and facial attributes from source to target candidate. Two of them were computer graphics-based and the other two utilizes a deep learning approach. It contains dataset in three types of compression factors raw, medium and high to make the models more robust towards detection. Celeb-DF dataset released later in the year 2019, contains 560 real videos and 5639 deepfake videos. Recently, Facebook has also hosted an online challenge on Kaggle, DeepFake Detection Challenge (DFDC), releasing 10,000 fake videos and 19,000 pristine videos.  

\subsection{Deepfakes Video Classification}
With evolving computational capacity and threats of deepfakes, several methods evolved to classify fake videos. Learning the mismatch between visual artifacts, head poses variation, using segmentation masks and training shallow and deep networks are some of the contemporary approaches that have made use of to detect manipulations.  Two-stream CNN recognizes the tampered faces by training a face classification network and a patch triplet network. The classification network uses LeNet architecture to train the model. Patch Triplet network helps the model to force the embeddings of the same type of images closer. In MesoNet\cite{mesonet}, shallow architectures with an inclusion of inception module learns the discriminative features from frames. On the other hand, recent works\cite{survey_2} have proved that deep architectures outperform the shallow networks by a large margin. HeadPose network identifies the tampering by measuring the distance between synthesized image head pose and the original image head pose. On estimating the 3D head pose from 2D coordinates system, the landmarks of manipulated faces are shifted from the original faces. Li and Lyu, in their work, detect peculiar artifacts that are introduced by warping steps. Convolutional Neural Networks (CNNs) are trained to detect the lack of consistency between the final and initial image after applying various transformations. Matern \textit{et al.} worked on visual artifacts to detect manipulated images. They showed that using facial attributes, there's a discrepancy between an original video and a manipulated one that is easily noticeable. However, they evaluated their approach mainly on DeepFakes and Face2Face type of manipulations. In the Multi-task learning approach, classification, segmentation, and reconstruction is performed altogether to boost classification accuracy. An encoder-decoder approach to learns to reconstruct the image and then final activation is used for classification. Capsule networks are designed to use less number of parameters to overcome the need for training millions of parameters in deep neural networks. Capsule Forensics uses a dynamic routing algorithm to generate an activation map where the face has been manipulated. As we can see in \cite{celebdf}, these approaches don’t generalize well on new and more challenging datasets. From \cite{faceforensics}, we can see many approaches \cite{bayar, cozzolino, xception, mesonet, rahmouni, steganalysis} perform well when the resolution of the video is high. As the video is compressed by a medium or high factor, the performance of these models drops significantly. From 95+\% accuracy, it drops 50-60\%, and in the case of NeuralTextures, accuracy goes down to 50\%, which means the network is unable to learn any features at such low resolutions and just randomly annotating the videos as fake or real.

\section{Methodology}
We use a Multitask Cascaded CNNs (MTCNN)\cite{mtcnn} to extract faces out of frames. Based on the success of detection of fake and real videos of XceptionNet architecture from FF++ paper \cite{faceforensics}, we started off with it for dataset video classification. To combat the classification in low-resolution videos, we further analyzed several methods using recurrent neural networks, convolutional 3D network, and, then finally metric learning approach. Our architecture and methods involved are discussed in the following subsections.

\subsection{MTCNN}
Crops out images using Proposal, Refine and Output Net. Proposal network detect faces across multiple resolutions, then, refine net suppress the overlapping boxes using nonmax suppression. Finally, output network gives the bounded face using five landmarks.
% It helps to crop out faces from images via three successive networks: Proposal Net, Refine Net, and, Output Net. Proposal network detect faces at multiple resolutions and gives bounding boxes regression vectors using the Euclidean loss between predicted and ground truth coordinates. This outputs several bounding boxes, to suppress the overlapping boxes nonmaximum suppression is used. Then, refine network further refines the number of bounding boxes incorporating bounding box regression and nonmax suppression. Then, finally, output network, gives the bounding box on faces using five landmarks. Output net uses facial landmarks regression to regress the coordinates further such that maximum focus is on the face. It can detect faces in high-resolution frames as well as low-resolution frames. In figure below, we can see that frame is hazy, even, then MTCNN crops out the image perfectly.

\begin{figure}[htbp]
\centering
\includegraphics[width=0.7\columnwidth]{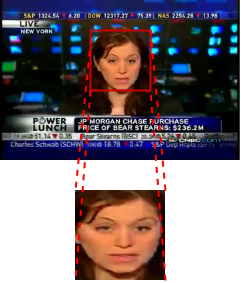} % Reduce the figure size so that it is slightly narrower than the column. Don't use precise values for figure width.This setup will avoid overfull boxes. 
\caption{Extraction of face from frame using MTCNN algorithm.}
\label{faceextraction}
\end{figure}

\subsection{Transfer Learning}
To make use of previous knowledge of architecture from one problem onto another problem is known as transfer learning. In our work, we used Xception\cite{xception} architecture to learn the crucial feature about real and fake faces. Xception net based on Inception V3 uses Inception module, with modification of the spatial convolutions to depthwise separable convolutions. After separating each channel, 1x1 depthwise convolutions helps network to capture the cross-channel correlations. Compared to Inception architecture convolutions, depthwise separable convolution differs in two ways: 1) Xception modules performs channel wise convolutions first, then, 1x1 convolution, compared to Inception where the 1x1 is performed earlier, and, 2) There's no non-linearity after depthwise separable convolutions. With this, the number of layers are reduced from 159 layers in Inception V3 to 126 layers in Xception architecture.

\subsection{Sequence Classification}
Recurrent Neural networks captures the information along the temporal domain. An output vector from the previous step is fed into the next step to learn the relation of features across time domain. Their ability to connect sequence of input frames over a period of time makes them significantly helpful for video classification purposes. LSTM supersedes the RNN as it can retain information for a long sequence of frames. In our work, we used a sequence of 16 and 32 frames per video to learn the inconsistency across the temporal domain. 

\subsection{3D Convolution}
3D convolution model employs 3D filters that pick up the knowledge of spatiotemporal features from the videos, in contrast to 2D convolution, where temporal domain is collapsed. In deepfakes, while transferring the appearance to target candidate, if the target candidate has a pose that's not happened in source candidate video then there's a discrepancy. To capture the spatial and temporal irregularities, we took 32 frames per video into consideration.

\subsection{Triplet Network}
Triplet network is a type of metric learning where the similar features are grouped together and different features are placed large apart in the feature space. 
The network applies loss to cluster the similar features together and difference amongst them in feature space. Let's take the anchor input sample A, a sample with the same label as P, and a sample with a different label as N. The loss function of triplet is defined as follows: 

\begin{equation}
\scriptstyle L(A, P, N) = \scriptstyle max(\left \| f(A) - f(P) \right \|^{2} - \left \| f(A) - f(N) \right \|^{2} + \alpha , 0) \\
\label{eq_2}
\end{equation}
where $\alpha$ is the margin (hyperparameter).

There are three different type of triplets generation methods based upon the distance between anchor, positive and negative embedding vectors. 
\begin{itemize}
    \item Easy Triplets: In this case, distance between negative and anchor embedding is greater than the distance between anchor and positive embedding plus  margin, i.e. $d(a,p) + margin<d(a,n)$. Hence, the loss propagated is zero and it does not help the network to learn anything.
    \item Semi-hard Triplets: Distance between anchor and negative is between the distance between anchor and positive, and, distance between anchor and positive plus margin, i.e.   $d(a,p)<d(a,n)<d(a,p)+margin$. The loss propagated is positive and zero in this scenario.
    \item Hard Triplets: The distance between anchor and negative is less than the distance between anchor and positive plus margin, i.e. $d(a,n)<d(a,p)+margin$ Hence, the loss propagated backwards is always positive in this case.
\end{itemize}

\subsection{Architecture}
For Celeb-DF dataset, we used Xception architecture trained end-to-end on the faces extracted via MTCNN. For FF++ high compression videos, we used semi-hard triplets to discriminate between the fake video and real video embedding vectors. MTCNN extract faces from the frames, then, facenet generates 512 dimension embeddings for each face in the feature space. As facenet is developed for face recognition, each unique face occupies a small cluster in the feature space. Then, we generate semi-hard triplets via online triplet mining. Using these triplets, the embeddings of fake frames and positive frames is distinctively separated through triplet loss. Embeddings is visualized in Figure \ref{tsne_plots_ff++}. To the best of our knowledge, use of metric learning for deepfakes video classification has not been explored yet. 

\begin{figure}[htbp]
\centering
\includegraphics[width=\columnwidth]{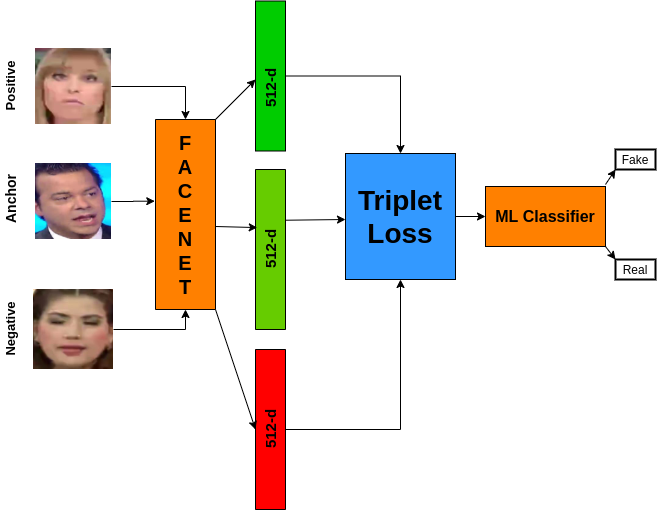} % Reduce the figure size so that it is slightly narrower than the column.
\caption{Triplet architecture used for clustering and classification of fake and real videos embeddings.}
\label{triplets}
\end{figure}

\section{Experiments Analysis}

\subsection{Datasets Review}
We analysed our video classification approaches on two datasets Celeb-DF and FF++ dataset. Training and testing set is provided by the dataset moderators.

\subsubsection{FF++ (c40)}
The dataset comprises four types of forgery videos namely: DeepFakes, Face2Face, FaceSwap and NeuralTextures. Each category contains 1000 videos taken from YouTube randomly. There are total 1000 pristine and 4000 forged videos. Frame size in raw, medium compressed and high compressed videos are 1080p, 720p and 480p respectively. Videos are captured in H.264 format, with compression factors of 0(raw), 23(medium) and 40(high). 

\subsubsection{Celeb-DF}
Celeb-DF comprises of 52 celebrities whose interviews are available on YouTube. They considered various factors such as gender, age and ethnic group bias to make the dataset more challenging. They created 5639 deepfake videos by swapping the faces  amongst 59 celebrities. The frame size is arbitrary in these videos. Video format is MPEG4.

\subsection{Implementation Details}
Initially, we used the frame rate of 5 for the Celeb-DF dataset to save frames from images. In case of FF++ dataset, we used the frame rate 1 to save all the frames. We analyzed the behaviour of our approach by increasing the number of frames from 10 frames per video till 25 frames per video. 

\subsubsection{Celeb-DF} 
Keeping the frame rate 5, number of negative frames extracted were approximately 66.5K frames and number of positive frames were 9.5k. Due to large data imbalance between fake and real videos, directly fine-tuning models pre-trained on imagenet data leads to poor performance. To counter the effect of data imbalance, we employed the bagging paradigm. Dividing the dataset into seven equal parts, we considered 1400 videos each time for training, totalling to approximately 19K frames. As the Imagenet dataset contains images from varied number of classes such as plant life, sports, animals, geological formation and person, we trained Xception end-to-end, with the face data, to make the lower layers focused on facial attributes. As we had enough number of frames to train, 22 million parameters were optimized after 30-50 epochs. With this approach, we produced 7 prediction outputs and took the maximum voting. We got the highest accuracy score of 99.8\% and AUC score of 99.2\%. 

\begin{figure}[htbp]
\centering
\includegraphics[width=0.8\columnwidth]{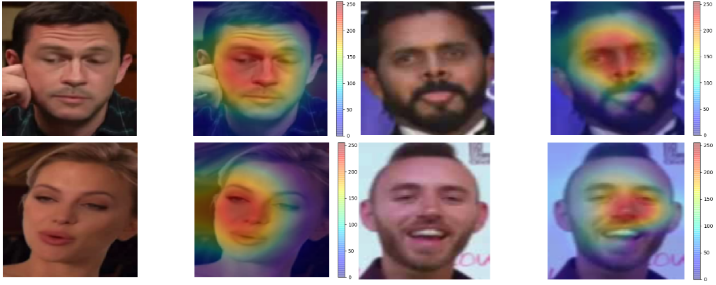} % Reduce the figure size so that it is slightly narrower than the column.
\caption{GradCAM\cite{gradcam} activation maps. From the above figure, we can see that the network focuses on the facial features helpful for binary classification.}
\label{gradcam}
\end{figure}

\begin{table}[htbp]
\begin{center}
\begin{tabular}[width=0.8\columnwidth]{c|c}
\hline
Method & AUC Score  \\
\hline
MesoInception4 & 53.6\\
Two-Stream & 53.8\\
Muti-task & 54.3\\
HeadPose & 54.6\\
Meso-4 & 54.8\\
VA-MLP & 55.0\\
VA-LogReg & 55.1\\
FWA & 56.9\\
Capsule & 57.5\\
DSP-FWA & 64.6\\
Xception & 65.5\\
Ours & \textbf{99.2} \\
\hline
\end{tabular}
\end{center}
\caption{AUC Scores on Celeb-DF dataset.}\label{tab1}
\end{table}

\begin{figure}[htbp]
\centering
\includegraphics[width=0.9\columnwidth]{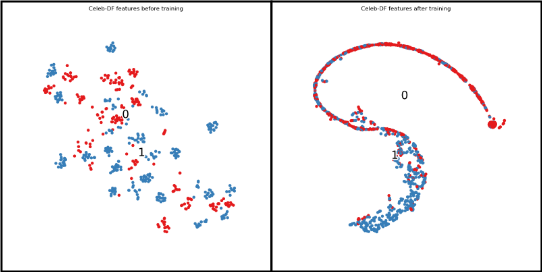} % Reduce the figure size so that it is slightly narrower than the column.
\caption{Celeb-DF features TSNE plot before and after training. \textit{Best viewed in color.}}
\label{tsne_celeb_df}
\end{figure}

\subsubsection{FF++}
From Rossler \textit{et al.} \cite{faceforensics} paper, we can see that  with raw and c23 compressed videos classification accuracy is almost around 99\% and 97\%, whereas in compression factor 40, the accuracy is below 90\%. Amongst all types of forgeries in the FF++ dataset, detecting NeuralTexture forgery is the most difficult as the accuracy goes down to below 80\%. To start with, like Celeb-DF we extracted frames with frame rate of five. We got around 29.5K number of frames. Amongst all imagenet models, XceptionNet outperformed other models. In   \cite{faceforensics}, the authors used 243K frames for training and 34K frames for validation. With a significantly reduced training data of 29K frames, we got accuracy of around 50\%, %i.e. that it's just classifying images randomly. 
%The network learns nothing from those images. 

After that, we added recurrence networks to make the network learn the sequences. It is with this idea that while transferring the source face on target face, there's information along the temporal dimension due to pose variation. From the trained XceptionNet on FF++ dataset, we generate embeddings from the lost pooling layer of feature vector dimension 5x5x2048. We also employed the LSTM architecture evaluation from dense layer features of 512 dimension. However, in this case too, the accuracy was around 55\% only. 

Generating features from trained end-to-end Xception was crucial, otherwise, the LSTM network overfits on the embedding data. It improved the accuracy by 2-3\%, but it was still below par.
To combine the features along spatial and temporal dimensions, we used 3D convolution. Without prior knowledge as in imagenet networks, conv 3d performs worse than even basic training of Xception on frames only. 

For triplet network, we initially took 29K frames and then generated a 512 dimension vector from trained Xception network. As the Xception network learnt nothing even after increasing the number of training frames to 43K, the embeddings generated did not provide us with any substantial gain while applying triplet loss to them. Then, we used FaceNet\cite{facenet} to generate face embeddings of dimension of 512. These face embeddings are clustered into n number of small groups in the feature space. Now, when we apply triplet loss to these embeddings, network learns the discriminative features to cluster the embeddings of original and manipulated faces separately. We analyzed triplets performance by randomly pick and online semi-hard negative triplet mining. 

\begin{figure*}[htbp]
\centering
\includegraphics[width=0.75\textwidth]{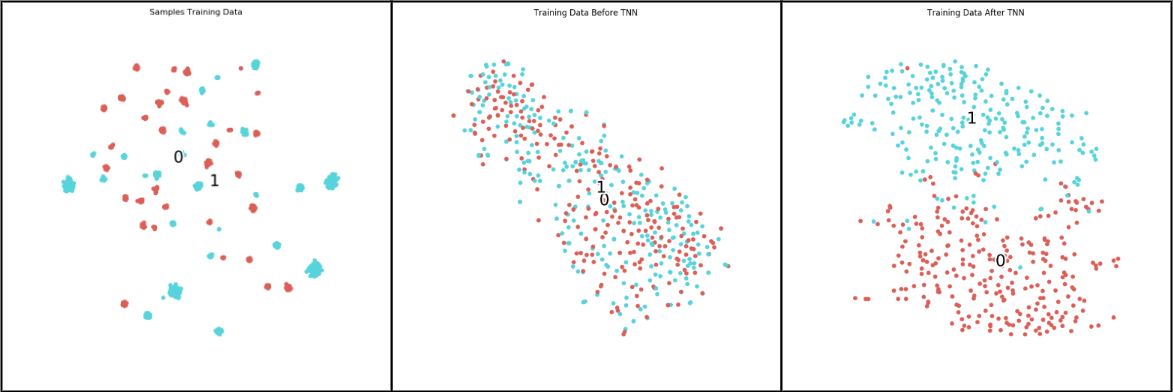} % Reduce the figure size so that it is slightly narrower than the column.
\caption{TSNE plots of FF++ dataset: a) Initial face embeddings vector generated from facenet architecture; b) Distribution of embeddings without triplet network; c) Distribution of embeddings in feature space after applying triplet loss. \textit{Best viewed in color.}}
\label{tsne_plots_ff++}
\end{figure*}
\begin{table}[t]
\begin{center}
\begin{tabular}[width=\textwidth]{c|c}
\hline
Model & Accuracy \\
\hline
Frames only & 49.82 \\
3d Convolution & 43.47 \\
Frames + LSTM/GRU & 55.8 \\
Triplets (Semi-hard) & \textbf{86.74}\\
\hline
\end{tabular}
\end{center}
\caption{FaceForensics++(c40) accuracy performance}\label{ff_accuracy}
\end{table}

\subsection{Performance Analysis}
In Celeb-DF, training the network with XceptionNet, we used Nadam\cite{nadam} optimizer with learning rate value of 0.002 and schedule decay of 0.004. We also tried with multi-loss function as mentioned in \cite{mesonet} but the performance gain was insignificant. From Table \ref{tab1}, we can see that using MTCNN for face extraction and training Xception Net on faces outperforms the AUC scores of all the previous approaches. With only 19K frames as compared to FF paper of using 240K frames, we acquired the accuracy of 96\%. By bagging and boosting algorithm, our accuracy got boosted by 2-3\% to 98\%. On FF++ dataset, we look into several methods for video classification. We reported the accuracy, recall, precision and F1 score in Table \ref{ff_accuracy} \& Table \ref{tab2}. In LSTM and C3D, we increased the number of sequence frames per video from 16 to 32, but, that does not help to distinguish between the authentic and tampered faces. Using triplet loss, however, provides us with sharp gain over traditional deep learning approaches.  From Fig. \ref{tsne_plots_ff++}, we can see that initially embeddings after facenet are clustered all over the feature space. Without training, just simple classification, the embeddings are mixed all together and are inseparable. After applying triplet loss, we have shown that there is a distinctive boundary between the fake videos and real videos. On top of triplet loss, we applied Random Forest (RF) and Stochastic Gradient Descent (SGD) to do binary classification of videos. We have shown an AUC and EER score of 92.9\% and 10.07\% that shows the robustness of our approach. When false rejection rate and false acceptance rate are equal, then we calculate the equal error rate. The  minimum the error, better the network.

\begin{table}[htbp]
\begin{center}
\begin{tabular}[width=0.9\textwidth]{c|ccccc}
\hline
Model & P & R & F1 & AUC & EER  \\
\hline
Triplets + RF & 89.84 & 82.73 & 86.14 & 92.85 & 12.14\\
Triplets + SGD & \textbf{90.55} & \textbf{82.74} & \textbf{86.47} & \textbf{92.9} & \textbf{10.07}\\
\hline
\end{tabular}
\end{center}
\caption{Models performance comparison using Semi-Hard Triplets. Here, P=Precision, R=Recall and EER = Equal Error Rate.}\label{tab2}
\end{table}

\begin{table}[htbp]
\begin{center}
\begin{tabular}[width=\columnwidth]{c|cc}
\hline
Model & NeuralTexture & Pristine  \\
\hline
Steg. Features + SVM \cite{steganalysis} & 55.84 & 56.94\\
Cozzolino \textit{et al.} \cite{cozzolino} & 62.15 & 56.27\\
Bayar and Stamm\cite{bayar} & 74.36 & 53.87\\
Rahmouni \textit{et al.}\cite{rahmouni} & 59.99 & 56.79\\
MesoNet\cite{mesonet} & 44.81 & 77.58\\
XceptionNet\cite{xception} & 78.06 & 75.27\\
Ours & \textbf{90.71} & \textbf{82.73}\\
\hline
\end{tabular}
\end{center}
\caption{Comparison of our approach to previous methods.}\label{tab3}
\end{table}

\begin{figure}[htbp]
\centering
\includegraphics[width=0.8\columnwidth]{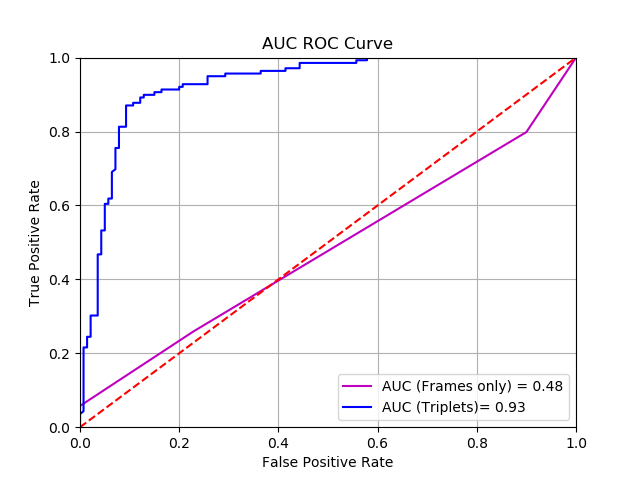} % Reduce the figure size so that it is slightly narrower than the column.
\caption{AUC ROC Curve plots of frames only and triplets network.}
\label{curve}
\end{figure}

\section{Conclusion and Future Work}
In this work, we presented a deep study for binary classification of deepfake videos. We analysed different approaches to improve the video classification in high compression factor using less amount of data. Using Triplet network, we outperform the previous results by a substantial margin utilizing only 25 frames per video. We also studied the effects of deep imagenet architectures on second-generation deepfakes dataset. Till now, the major limitation of the approaches is their generalizability across different datasets. In future, our aim is to use unsupervised domain adaptation to adapt the feature space from source dataset to target dataset, to make our model robust and label independent.

{\small
\bibliographystyle{IEEEtran}
\bibliography{ref}
}

\end{document}